%% file: main.tex
\documentclass[10pt,twocolumn,letterpaper]{article}
\usepackage[accsupp]{axessibility}
\usepackage{cvpr}

\definecolor{cvprblue}{rgb}{0.21,0.49,0.74}
\usepackage[pagebackref,breaklinks,colorlinks,allcolors=cvprblue]{hyperref}

\title{Self-Supervised Pretraining for Fine-Grained Plankton Recognition}

\author{
    Joona Kareinen\textsuperscript{1}, Tuomas Eerola\textsuperscript{1}, Kaisa Kraft\textsuperscript{2}, Lasse Lensu\textsuperscript{1}, Sanna Suikkanen\textsuperscript{2}, Heikki K\"alvi\"ainen\textsuperscript{1,3} \\
    \textsuperscript{1}LUT University, Computer Vision and Pattern Recognition Laboratory, Lappeenranta, Finland\\
    \textsuperscript{2}Finnish Environment Institute, Helsinki, Finland\\
    \textsuperscript{3}Brno University of Technology, Faculty of Information Technology, Brno, Czech Republic
}

\begin{document}
\maketitle
\input{sec/0_abstract}    
\input{sec/1_introduction}

\input{sec/2_related}

\input{sec/3_methods}

\input{sec/4_experiments}
\input{sec/5_conclusion}

{
    \small
    \bibliographystyle{ieeenat_fullname}
    \bibliography{main}
}

\input{sec/X_suppl}

\end{document}

%% file: sec/0_abstract.tex
\begin{abstract}
Plankton recognition is an important computer vision problem due to plankton's essential role in ocean food webs and carbon capture, highlighting the need for species-level monitoring. However, this task is challenging due to its fine-grained nature and dataset shifts caused by different imaging instruments and varying species distributions. As new plankton image datasets are collected at an increasing pace, there is a need for general plankton recognition models that require minimal expert effort for data labeling. In this work, we study large-scale self-supervised pretraining for fine-grained plankton recognition. We first employ masked autoencoding and a large volume of diverse plankton image data to pretrain a general-purpose plankton image encoder. Then, we utilize fine-tuning to obtain accurate plankton recognition models for new datasets with a very limited number of labeled training images. Our experiments show that self-supervised pretraining with diverse plankton data clearly increases plankton recognition accuracy compared to standard ImageNet pretraining when the amount of training data is limited. Moreover, the accuracy can be further improved when unlabeled target data is available and utilized during the pretraining.

\end{abstract}

%% file: sec/1_introduction.tex
\section{Introduction}
\label{sec:intro}
Despite the advancements in vision foundation models, fine-grained recognition in the presence of dataset shift remains a challenging task~\cite{stevens2024bioclip}. The subtle differences between classes, combined with distribution shifts in class appearance and composition between training and target datasets, necessitate image representations that are both general enough to handle dataset shifts and specific enough to distinguish visually similar classes. These challenges are difficult to tackle with a general-purpose vision foundation model and call for tailored solutions. Plankton recognition offers an interesting and environmentally relevant application for studying and developing such methods. Recognizing taxonomically close plankton species is challenging and is further complicated by domain and class distribution shifts across the existing datasets.

\begin{figure}[t]
    \centering
    \includegraphics[width=\linewidth]{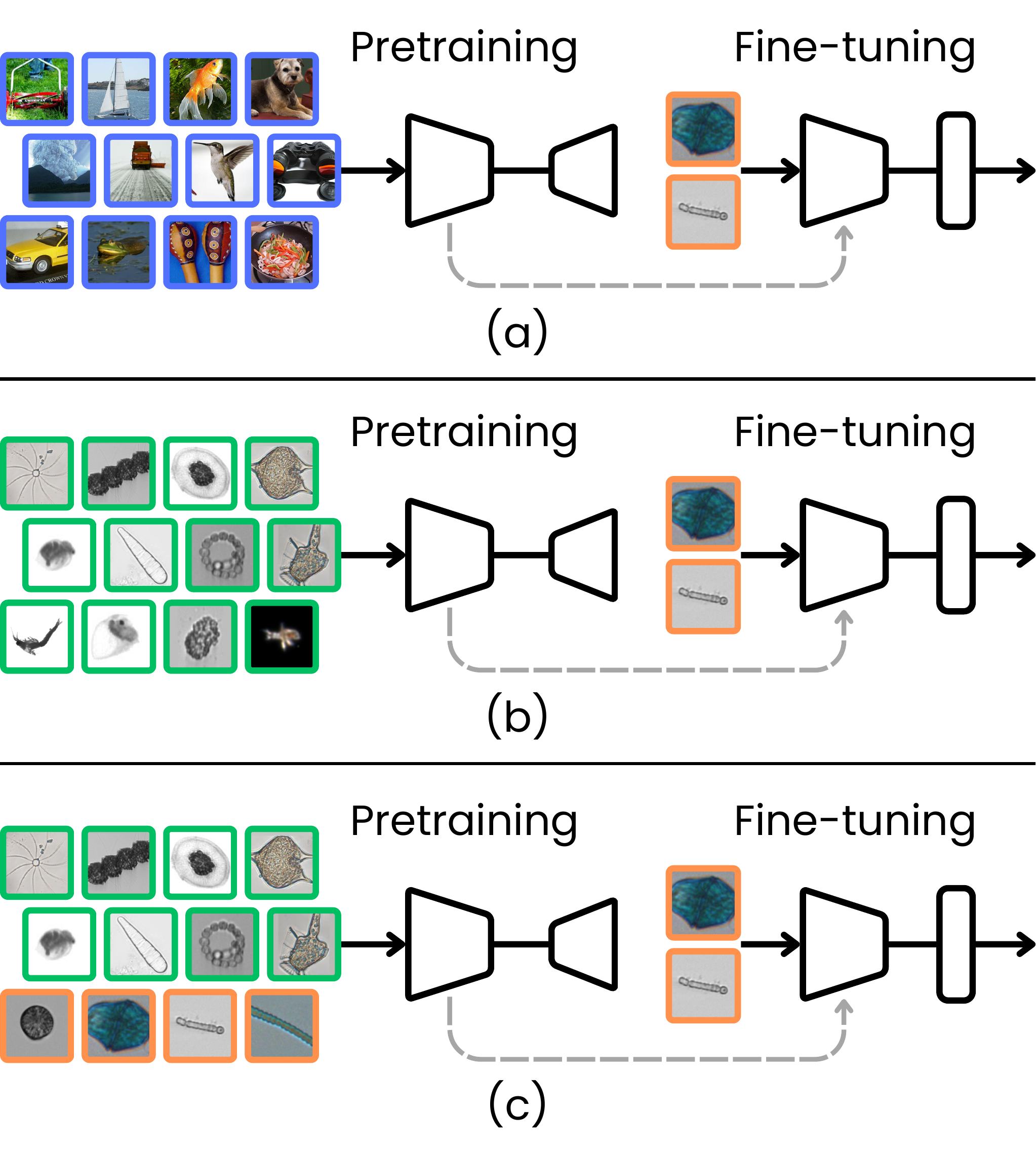}
    \caption{We evaluate three different self-supervised pretraining strategies: (a) pretraining on ImageNet-1k and fine-tuning on plankton data, (b) pretraining on a diverse plankton dataset and fine-tuning on unseen plankton data, and (c) pretraining on a diverse plankton dataset and fine-tuning on a subset of the same dataset.}
    \label{fig:summary}
\end{figure}

Plankton are a collection of microscopic organisms that serve as the foundation of aquatic ecosystems. These organisms are broadly classified into two primary groups: phytoplankton and zooplankton. Phytoplankton are critical as primary producers at the base of the food chain, contributing approximately 50\% of the world's oxygen production and around 40\% of global carbon fixation~\cite{falkowski1994role, field1998primary}. Similarly, zooplankton are critical in matter and nutrient cycling from the primary producers to higher trophic levels, such as commercially significant fish species~\cite{beaugrand2003plankton}. Furthermore, zooplankton transport carbon from surface to deep waters through feeding, daily migrations, and fecal pellets, playing a key role in global carbon cycles~\cite{Banse1995, Turner2015}. 

In addition to being vital to the marine food web, plankton also serve as an indicator of ocean health. Plankton are sensitive to environmental changes: they respond rapidly to changes in temperature, nutrient availability, and water flows~\cite{hays2005}. This makes them valuable for understanding the aquatic ecosystem dynamics, which in turn contributes to the prediction of environmental changes such as pollution and climate change~\cite{hays2005}.

Understanding plankton species distributions provides valuable information on the oceans' health, but monitoring plankton communities and composition requires sustained observations and is challenging due to their diversity and dynamic changes. To address this, various plankton imaging and analysis instruments have been developed for both \textit{in-situ} and laboratory use~\cite{global2019}. Recent technological advancements have led to the development of both automated and semi-automated imaging systems that can be used to capture massive plankton datasets. However, the use of plankton imagery faces various challenges, such as the fine-grained nature of the identification task, the rarity of species, and the uncertainty of category labeling~\cite{eerola2024survey}. There is a growing need for automated plankton species identification systems, as they significantly reduce the need for manual identification and improve the efficiency of processing large amounts of data.

As plankton imaging instruments become more accessible, an increasing number of plankton image datasets are being collected. These datasets vary in both the imaging instruments used and the plankton species they contain. This causes dataset shifts between them, and therefore, a plankton recognition model trained on one dataset often fails when applied to another~\cite{batrakhanov2024daplankton}. Manually labeling sufficient training data for each dataset to create dataset-specific models is impractical. Therefore, a more general model that minimizes the need for expert labeling would be highly beneficial. However, training a single general model using multiple datasets in a supervised manner is challenging due to the lack of a universally agreed-upon method for categorizing plankton, although connecting the classes to the World Register of Marine Species is routinely done by many users. The set of class labels in a plankton dataset depends on various factors, such as the environmental application for which the dataset was collected, the range of particle sizes the imaging instrument can capture, and the geographic region from which the water samples were collected. Moreover, the full taxonomical hierarchy of plankton is undergoing continuous revision as recent studies have deepened the understanding of their biodiversity~\cite{simon2009diversity}. Therefore, the class labels between datasets are not always entirely comparable, making it difficult to use them simultaneously for supervised training. This issue is especially notable in classes that do not correspond to a specific species, as they are particularly affected by taxonomic revisions. 

In this paper, we address the issue of incomparable class labels between datasets through self-supervised pretraining. First, we collect a large pool of public plankton image datasets and use unsupervised learning to obtain general-purpose plankton representations without the need to harmonize class labels across datasets. For this, we employ Masked Autoencoders (MAE)~\cite{he2022masked}, which guide the encoder to learn plankton image representations by masking a large portion of input images and then learning to generate the missing parts. This masking technique allows the model to learn features that effectively describe the subtle visual differences between fine-grained plankton classes by forcing it to focus on small regions in the images. Next, we fine-tune the encoder with an incorporated classification head on plankton datasets with a limited amount of labeled training images.

In the experimental part of the work, we compare different pretraining methods on plankton data, as shown in \Cref{fig:summary}. We evaluate our models pretrained on plankton image data against standard ImageNet pretraining in varying labeled data scenarios. Our results show that domain-specific pretraining can obtain results comparable to ImageNet pretraining with significantly fewer images. Additionally, we show that domain-specific pretraining can surpass ImageNet pretraining in accuracy, particularly when labeled data is limited.

The main contributions of this study are as follows: (1) The first application of Masked Autoencoders for self-supervised pretraining on plankton image data, (2) Extensive evaluation of different pretraining strategies under varying labeled data conditions, and (3) Analysis of the benefits of domain-specific pretraining on fine-grained plankton recognition.

%% file: sec/2_related.tex
\section{Related works}
\label{sec:related_works}

\subsection{Plankton recognition}

A large number of plankton recognition methods have been proposed, starting from traditional feature engineering methods to more modern deep learning methods~\cite{eerola2024survey}. Early plankton recognition methods relied on hand-picked features such as shape, texture, and size to obtain representative feature vectors, which were then classified using methods such as support vector machine or decision trees. 
In the past decade, convolutional neural networks (CNNs) have become the dominant method for plankton recognition. Plenty of different CNN architectures have been used for plankton recognition, including custom ones~\cite{deepsea2015,lumini2019deep,burevs2021plankton,kraft2022towards}.

More recently, vision transformers (ViTs)~\cite{Kolesnikov2021vit} have been shown to outperform CNNs on various plankton datasets~\cite{maracani2023domain, callejas2025}. ViTs work by splitting the image into patches and converting these patches into vectors, which are then processed using self-attention mechanisms. The self-attention allows the model to capture relationships between different parts of the image, enabling it to identify the most informative regions. Kyathanahally \etal\cite{kyathanahally2022ensembles} tested ensembling multiple Data-efficient image Transformers (DeiTs)~\cite{touvron2021deit} for various ecological datasets, including four plankton datasets. Maracani \etal~\cite{maracani2023domain} applied in-domain and out-of-domain transfer learning in plankton recognition. Notably, they demonstrate that a model pretrained on ImageNet-21K and then fine-tuned on plankton data achieves better results compared to a model trained from scratch using a single plankton dataset. Callejas \etal~\cite{callejas2025} tested different ViTs and CNNs in the classification of ZooScanNet~\cite{zooscannet} and WHOI-Plankton \cite{WHOIplankton}, the two largest publicly available plankton datasets, and reported that ViTs achieve the best accuracies. An extensive review of existing plankton recognition methods can be found in~\cite{eerola2024survey}.

The vast majority of existing plankton recognition models have been trained and evaluated on one or a few, often in-house, datasets. Due to dataset shifts caused by different imaging instruments and varying class compositions, these models do not generalize well to previously unseen plankton datasets. Additionally, efforts have been made to develop more general plankton recognition methods using transfer learning~\cite{orenstein2017transfer,maracani2023domain}, domain adaptation~\cite{batrakhanov2024daplankton}, and open-set recognition~\cite{pu2021anomaly,badreldeen2022open,kareinen2024open}.

\subsection{Self-supervised learning}
Self-supervised learning (SSL) refers to a family of techniques that enable models to learn meaningful representations from unlabeled data~\cite{jaiswal2020survey}. By reducing the dependency on manually annotated datasets, SSL improves data efficiency, especially in domains where labeled data is scarce or expensive to obtain.  

Some of the early deep SSL methods in computer vision were  Generative Adversarial Networks (GANs)~\cite{goodfellow2014generative} and autoencoders~\cite{schmidhuber2015deep}. GANs consist of a generator that creates realistic-looking images and a discriminator that tries to detect generated images from real samples. Through this training process, GANs indirectly capture meaningful image representations but are primarily designed for image generation rather than representation learning. Autoencoders, on the other hand, encode input images into a compact latent representation and then reconstruct them from this latent space. While both GANs and autoencoders showcased the potential of learning from unlabeled data, they focused on generative tasks rather than learning transferable representations for downstream tasks.

A major advancement in SSL was contrastive learning, in which a model learns representations by grouping similar (positive) samples closer while pushing dissimilar (negative) ones apart in the feature space. In the absence of labels, positive and negative pairs are typically defined using data augmentation: different augmented views of the same image are treated as positive, while all other images are considered negative. Unlike the generative approaches, contrastive learning directly optimizes for feature discrimination, making it well-suited for transfer learning.

Momentum Contrast (MoCo), proposed by He \etal~\cite{he2020momentum}, introduced a memory queue that stores a large number of negative samples across training batches, addressing the batch size limitations in contrastive learning. MoCo utilizes a momentum-updated encoder, which prevents representation collapse by updating a secondary encoder using an exponential moving average of the main encoder's parameters. A simple framework for contrastive learning (SimCLR)~\cite{chen2020simple} simplified contrastive learning by removing the need for queue samples and instead relied on a large batch size to generate a diverse set of negative samples. SimCLR applies two random augmentations to each data sample, passing them through a shared encoder and projection head, and maximizes the similarity between the augmented views. After self-supervised training, the projection head is removed, and only the encoder is used for downstream tasks. SimCLR demonstrated that strong data augmentations in SSL significantly improve the quality of learned representations. Barlow Twins~\cite{zbontar2021barlow} is a novel approach to contrastive learning through redundancy reduction. Instead of using contrastive loss, Barlow Twins minimize the off-diagonal elements in a cross-correlation matrix between representations of two augmented views, effectively encouraging representations to be invariant to distortions. 

Bootstrap Your Own Latent (BYOL)~\cite{grill2020bootstrap} further advanced SSL by showcasing that it can achieve state-of-the-art results without any negative pairs. Later, self-distillation with no labels (DINO)~\cite{caron2021dino} introduced a self-distillation framework inspired by BYOL. DINO trains a student network using a cross-entropy loss between the student and a momentum-updated teacher network's predictions. DINO demonstrated that ViTs can learn strong representations through self-supervised training without requiring large labeled datasets. Further improvements to this method were introduced in DINO v2~\cite{dinov2}.

Masked image modeling (MIM) has emerged as a recent approach to SSL, inspired by masked language modeling in natural language processing (\eg BERT~\cite{devlin2019bert}). Instead of comparing augmented image pairs, MIM methods train a model to reconstruct missing parts of an image. BEIT~\cite{bao2021beit} is a token-based masked model where the image is first tokenized into discrete visual embeddings, and the model learns to predict masked tokens. This approach was further improved in BEIT v2~\cite{beitv2}. Masked Autoencoder~\cite{he2022masked} represents a more efficient patch-based masking strategy, reconstructing raw pixel patches using a lightweight decoder. SimMIM~\cite{xie2022simmim} simplified the approach even further by applying a direct pixel-wise loss. 

\subsection{Semi-supervised learning in fine-grained recognition}

Semi-supervised learning approaches have evolved significantly in recent years. Traditional methods typically utilize both labeled and unlabeled data during the training process, whereas another approach is to first utilize self-supervised learning followed by supervised fine-tuning. The latter strategy has become more popular after Chen \etal~\cite{chen2020big} demonstrated that large self-supervised models excel when fine-tuned with small amounts of labeled data.

Su \etal~\cite{su2021realistic} evaluated the existing semi-supervised learning methods for fine-grained image classification. The results revealed that semi-supervised learning methods can improve performance compared to training from scratch. Their research further showed that self-supervised pretraining can benefit significantly from out-of-distribution data. Kim \etal~\cite{kim2023coreset} introduced a novel open-set self-supervised learning in which the pretraining dataset contains instances of relevant or irrelevant domains to the target dataset. They proposed the SimCore algorithm that samples a subset of the open-set that is semantically similar to the target dataset and demonstrated enhanced representation learning performance in fine-grained recognition tasks. Duan \etal~\cite{duan2024roll} proposed a pseudo-label selection method that encouraged pseudo labels to include likely ground truth labels while excluding noisy ones.

In the domain of plankton recognition, researchers have explored various self-supervised and semi-supervised learning methods that typically utilize clustering to allow the potential new-class discovery from unlabeled data. Early work by Salvesen \etal~\cite{salvesen2020robust} introduced an autoencoder-based framework with an embedded clustering layer that learns a latent representation while simultaneously performing unsupervised clustering. Later, they extended the approach with rotation invariant features and spectral clustering to refine class separation~\cite{salvesen2022robust}. Schmarje \etal~\cite{schmarje2021fuzzy} proposed a semi-supervised learning framework for handling fuzzy labels, where a small set of certain images is used to guide clustering of a large set of uncertain fuzzy images. Schröder \etal~\cite{schroder2022assessing} compared multiple feature extraction methods under varying assumptions of label and data availability and tested two new semi-supervised learning methods tailored for MorphoCluster~\cite{schroder2020morphocluster}.

Recent advances have focused more on enhancing feature representation with
Schanz \etal~\cite{schanz2023robust} applying SimCLR pretraining on CNN models while addressing the issue with imbalanced classes. Pastore \etal~\cite{pastore2023efficient} proposed an unsupervised learning method that combined features from multiple pretrained CNNs and ViTs and further compresses them to be used in clustering. In \cite{oldenburg2023deeploki}, the authors compared both transfer learning and DINO pretraining with CNNs and reported that DINO pretraining obtained better results.

%% file: sec/3_methods.tex
\section{Methods}
\label{sec:methods}

\subsection{Masked autoencoder}

Masked autoencoder (MAE) is a self-supervised learning method proposed by He \etal \cite{he2022masked}. MAE is designed to reconstruct missing portions of an input image by leveraging only small unmasked parts of the information. Like standard autoencoders, MAE consists of an encoder that maps the input image into a latent representation and a decoder that reconstructs the original input from this representation. However, unlike traditional autoencoder architectures, MAE uses asymmetric design where the encoder is significantly larger than the decoder. Recently, MAE has been shown to scale not only with model size but also with the size of the training dataset in larger models~\cite{singh2023effectiveness}.

MAE follows a patch-based processing approach similar to ViTs, where the input image is divided into non-overlapping patches. During training, a large random subset of these patches is masked. The encoder is applied only to the unmasked patches, which significantly reduces the computational complexity and allows the training of larger encoders. The masking encourages the encoder to learn meaningful representations from the unmasked patches that are useful for reconstruction. After the visible patches have been encoded, the decoder reconstructs the masked patches using the learned latent representations. 

The training objective in MAE is to minimize the reconstruction error between the original and the predicted masked patches. The loss function is formulated as 
\begin{equation}
    \mathcal{L} = \frac{1}{N_m} \sum_{i \in M} (x_i - \hat{x}_i)^2,
\end{equation}
where $N_m$ is the number of masked patches, $M$ is the set of masked patch indices, $x_i$ is an original patch, and $\hat{x}_i$ is its reconstruction. This loss function, based on mean squared error (MSE), is computed only for the masked patches and not for the visible ones. In the original paper \cite{he2022masked}, the authors showed that a high masking ratio is required for the model to learn transferable feature representations. For downstream tasks, the decoder is removed from the network, and only the encoder is used as a feature extractor. This architecture allows fine-tuning the model for specific recognition tasks while leveraging the feature representations learned during self-supervised pretraining.

%% file: sec/4_experiments.tex
\section{Experiments}
\label{sec:experiments}

\begin{table*}[htb]
    \centering
    \caption{Summary of the datasets used for pretraining.}
    \resizebox{\linewidth}{!}{%
    \begin{tabular}{llllrr}
        Dataset &  Imaging instrument(s) & Region & Plankton type & \# of species & \# of images \\ 
        \midrule[1.25pt]
        Kaggle-Plankton~\cite{Cowen2015}  & ISIIS-2 & Straits of Florida, U.S & zooplankton & 121 & 130,000 \\
        Lake Zooplankton~\cite{kyathanahally2021deep} & DSPC & Lake Greifensee, Switzerland & zooplankton & 35 & 18,000 \\
        SYKE-Plankton-ZooScan\_2024~\cite{zooscan2024} & ZooScan & Baltic Sea & zooplankton & 20 & 24,000 \\
        PMID2019~\cite{li2020developing} & Bright-field microscope & Jiaozhou Bay, China & phytoplankton & 24 & 14,000 \\
        SYKE-Plankton-IFCB\_2022~\cite{syke2022} & IFCB & Baltic Sea & phytoplankton & 50 & 63,000 \\
        UDE Diatoms in the Wild 2024~\cite{Kloster2024} & Bright-field microscope & -- & phytoplankton & 611 & 84,000 \\
        DAPlankton~\cite{batrakhanov2024daplankton} & IFCB, CS, FlowCam & Baltic Sea & phytoplankton & 44 & 112,000 \\ 
        \cmidrule{1-6}
        Total & & & & & 443,000 \\
    \end{tabular}}
    \label{tab:summary}
\end{table*}

\subsection{Data}

To obtain a diverse dataset for self-supervised pretraining, we combined data from seven publicly available plankton datasets. These datasets include Kaggle-Plankton \cite{Cowen2015}, Lake Zooplankton \cite{kyathanahally2021deep}, SYKE-Plankton-ZooScan\_2024 \cite{zooscan2024}, PMID2019 \cite{li2020developing}, SYKE-Plankton-IFCB\_2022 \cite{syke2022}, UDE Diatoms in the Wild 2024 \cite{Kloster2024}, and DAPlankton \cite{batrakhanov2024daplankton}. Together, these datasets contain both phyto- and zooplankton images and capture a wide range of species and imaging conditions. Class overlap between datasets is minimal, with the notable exception of DAPlankton and SYKE-Plankton-IFCB\_2022, which share 33 taxa (Full class overlap in supplementary material). A summary of these datasets is presented in \Cref{tab:summary}.

For Kaggle-Plankton, which contains a labeled training set with 30,000 images and an unlabeled test set with 130,000 images, we utilize only the larger test set for pretraining. DAPlankton contains two subsets: DAPlankton$_{\textrm{LAB}}$ and DAPlankton$_{\textrm{SEA}}$. Both subsets were used in the pretraining dataset. 

\begin{figure*}[htb]
    \centering
    \includegraphics[width=\linewidth]{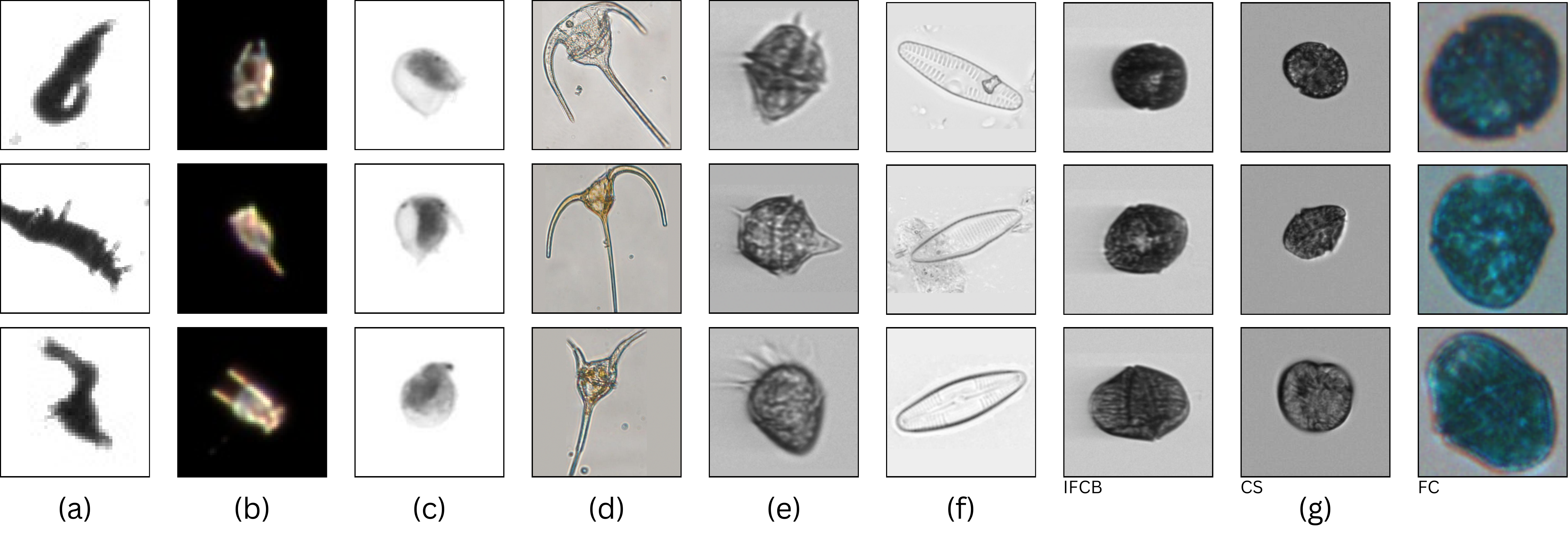}
    \caption{Example plankton images from different datasets: a) Kaggle-Plankton~\cite{Cowen2015}, b) Lake Zooplankton~\cite{kyathanahally2021deep}, c) SYKE-Plankton-ZooScan\_2024~\cite{zooscan2024}, d) PMID2019~\cite{li2020developing}, e) SYKE-Plankton-IFCB\_2022~\cite{syke2022}, f) UDE Diatoms in the Wild 2024~\cite{Kloster2024}, g) DAPlankton~\cite{batrakhanov2024daplankton}. The shown images are taken from three different, visually similar classes within each dataset, highlighting the fine-grained nature of the data. For DAPlankton, the same three classes are shown across all instruments.}
    \label{fig:plankton_img}
\end{figure*}

Our preprocessing strategy follows a similar approach to earlier works~\cite{lumini2019, maracani2023domain}. To maintain the original aspect ratio while creating a square image, we pad the smaller dimension using a background color that matches the image. This background color matching is done by 1) computing the mode color from all edges of the image and 2) estimating the background noise by calculating the standard deviation from 20\% of the edge pixels closest to the mode color. Dataset-specific preprocessing includes removing size indication legends from SYKE-Plankton-ZooScan\_2024 images and utilizing the supplied ground truth bounding box information to crop individual samples from PMID2019. Example images from each dataset after preprocessing are shown in \Cref{fig:plankton_img}. 

For fine-tuning and testing, we utilized both subsets of DAPlankton: DAPlankton$_{\textrm{LAB}}$ and DAPlankton$_{\textrm{SEA}}$. DAPlankton$_{\textrm{LAB}}$ consists of 47,471 images from 15 phytoplankton species captured with three different imaging instruments: Imaging FlowCytobot (IFCB)~\cite{olson2007submersible}, CytoSense (CS)~\cite{dubelaar1999design}, and FlowCam (FC)~\cite{sieracki1998imaging}. DAPlankton$_{\textrm{SEA}}$ consists of 64,453 images from 31 phytoplankton species captured using IFCB and CS. The DAPlankton$_{\textrm{SEA}}$ is the more realistic of the two datasets as the images were collected \textit{in-situ} from the Baltic Sea and the dataset is heavily imbalanced. The data compositions and the range of images per class are shown in \Cref{tab:daplankton_summary}.

\begin{table}[h]
    \centering
    \caption{Summary of the DAPlankton subsets.}
    \resizebox{\columnwidth}{!}{%
    \begin{tabular}{l c c c}
        \textbf{Dataset} & Instrument & \# of images & \# of images per class \\ 
        \midrule[1.25pt]
        DAPlankton$_{\textrm{LAB}}$ & IFCB  & 16,476  & 1,001 -- 1,376  \\
            & CS    & 13,187  & 608 -- 1,285    \\
            & FC    & 17,808  & 286 -- 2,618    \\
        \cmidrule{1-4}
        DAPlankton$_{\textrm{SEA}}$ & IFCB  & 51,622   & 54 -- 12,280  \\
            & CS    & 12,831 & 5 -- 5,443    \\
    \end{tabular}}
    \label{tab:daplankton_summary}
\end{table}

The DAPlankton dataset is unique among the publicly available plankton datasets as it contains the same shared classes across both partitions. Additionally, the laboratory partition has negligible label uncertainty as the cultures were grown in a laboratory and checked for cross-contamination. The shared taxonomy between different imaging instruments allows the evaluation of how different imaging instruments affect the model's performance on identical taxonomic targets.

\subsection{Design of experiments}
\subsubsection{Self-supervised pretraining}
Our implementation of MAE was built using PyTorch, PyTorch Lightning, and LightlySSL. 
The ViT models were implemented using the PyTorch image models (timm) library~\cite{rw2019timm}. 

\textbf{Image Augmentations}: The pretraining includes a series of image augmentations that make the pretraining dataset more versatile. The input images were resized to 256 $\times$ 256 pixels and converted to grayscale. The grayscale conversion was added to unify the plankton images as in any case, the color information in them is very limited. A random patch was selected with a scale ranging from  0.4 and 1.0 and resized to 224 $ \times $ 224 pixels. Following this, we applied random horizontal and vertical flips and normalization using the dataset mean and standard deviation. 

\textbf{Architecture}: We employed a ViT-L as our encoder backbone, following the original implementation from \cite{he2022masked}. The decoder was implemented as a lighter transformer architecture with 8 layers and an embedding dimension of 512. During pretraining, we randomly masked 75\% of the image patches, which was shown to be effective in the original MAE approach. The patch size was set to 16 $\times$ 16, resulting in 196 patches per image.

\textbf{Optimization}: We utilized AdamW optimizer~\cite{loshchilov2018decoupled} with $\beta_1 = 0.9$ and $\beta_2 = 0.95$. The learning rate followed the cosine decay schedule with linear warmup, where the base learning rate was set to 1.5e-4$\times$ (eff\_batchsize/256) following the linear scaling law~\cite{goyal2017accurate} with the weight decay set to $0.05$. The reconstruction loss was computed between normalized pixel values between the reconstructed and the original masked patches. 

\begin{figure}[htb]
    \centering
    \includegraphics[width=\linewidth]{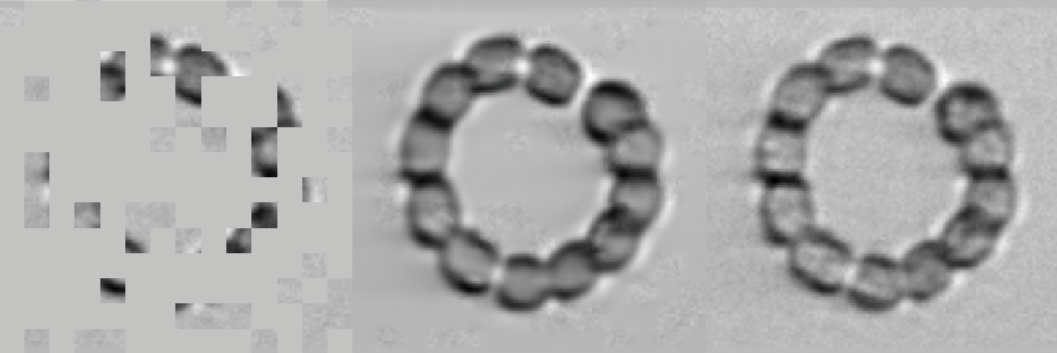}
    \caption{Masked autoencoder learns from the small unmasked patches (left) to reconstruct (middle) the original plankton image (right).}
    \label{fig:recon}
\end{figure}

We pretrained two different models from scratch: one using the complete pretraining dataset and another excluding the DAPlankton dataset and all classes present in both DAPlankton and SYKE-Plankton-IFCB\_2022 datasets. This reduction resulted in a pretraining dataset of 280,000 images. The first model, referred to as ViT-L (with-daplankton), was trained for 800 epochs using 8 nodes with 4 NVIDIA V100 GPUs per node, while the second model, ViT-L (no-daplankton), was trained using only 4 nodes. Both models used an effective batch size of 4,096, with the smaller training setup utilizing gradient accumulation to match this batch size. The results showcased high-quality reconstruction, as shown in \Cref{fig:recon}.

\subsubsection{Fine-tuning}

\textbf{Dataset and evaluation}:
We fine-tuned our pretrained models using both DAPlankton$_{\textrm{LAB}}$ and DAPlankton$_{\textrm{SEA}}$ datasets. Additionally, we utilized ViT-L model trained with the ImageNet-1k dataset as a transfer learning baseline. Each experiment ran for 50 epochs, and we applied 5-fold cross-validation. For each fold, the dataset was first split into 20\% testing and 80\% training, of which 15\% was selected as a validation set. For splitting the dataset, we utilized a stratified strategy, which ensured that each fold was similarly balanced. We applied the same set of augmentations for the training set as in pretraining, but for validation and testing, the vertical and horizontal flips were removed, and the crop was centered on the image. We evaluate our results by using accuracy as the main metric.

\begin{figure*}[htb]
    \centering
    \includegraphics[width=\linewidth]{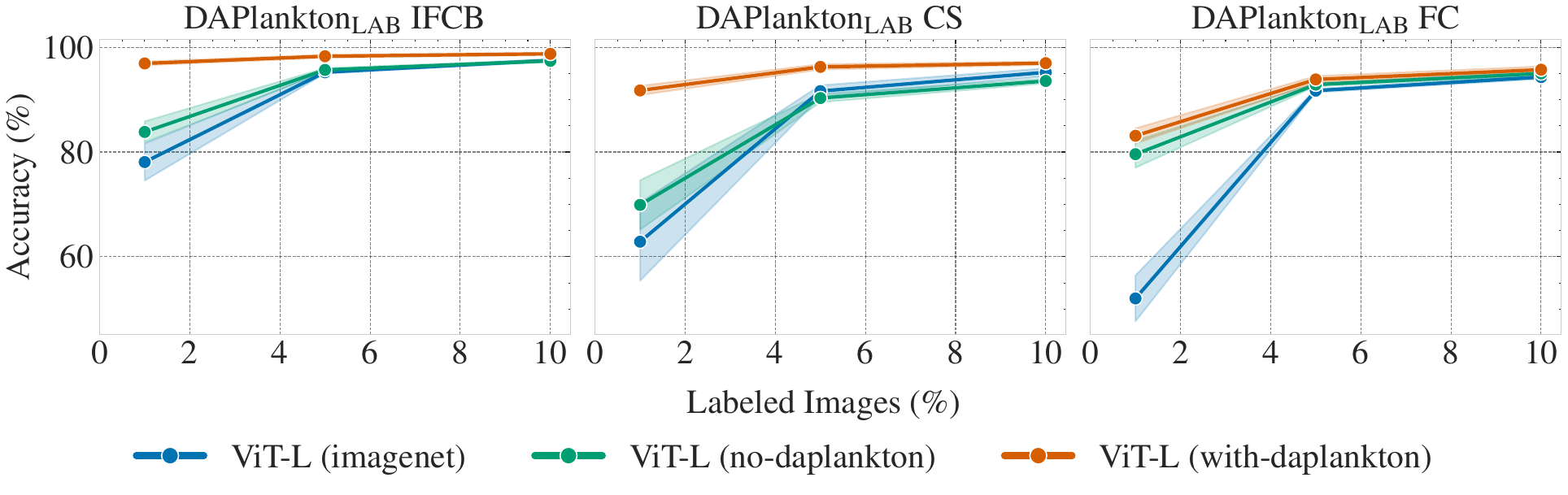}
    \caption{Mean accuracy and standard deviation for DAPlankton$_{\textrm{LAB}}$ across different labeled data subsets.}
    \label{fig:ft_lab}
\end{figure*}

\textbf{Model architecture}:
We adopted the bottleneck architecture from~\cite{maracani2023domain}, where the encoder's output is passed through a hidden layer of size 512, with LayerNorm and GELU activation \cite{GELU2016}, followed by the final classification layer. We utilized layer-wise learning rate decay \cite{bao2021beit} with a decay factor of 0.75 to gradually increase learning rates from deeper to shallower layers to preserve the pretrained features.

\textbf{Optimization}:
During fine-tuning, we set the learning rate to 2e-3 and used the AdamW optimizer with a weight decay of $0.01$. The learning rate followed a cosine decay with a 5-epoch warmup period.  As a loss function, we use cross-entropy loss with label smoothing~\cite{szegedy2016rethinking} of $0.1$. For further regularization, we applied stochastic depth \cite{huang2016deep} with a drop path rate of $0.2$, following the fine-tuning strategy from \cite{he2022masked}. We used a batch size of 128 and fine-tuned the models using a single NVIDIA A100 GPU.

\subsection{Results}
The results for fine-tuning using the full DAPLankton$_{\textrm{LAB}}$ dataset are presented in \Cref{tab:result_full}. The table shows the accuracies for all three pretrained models for all three partitions of the dataset. The results show that when using a large amount of training data for fine-tuning, there are no significant differences in accuracy between the pretraining approaches.
This indicates that pretraining a ViT-L with a diverse plankton dataset does not provide a notable benefit compared to a model pretrained with ImageNet-1K, when enough labeled images in the target dataset are available. However, it should be noted that ImageNet-1K is 3-4 times larger than the used combinations of plankton datasets, implying that a smaller amount of data is sufficient for pretraining if it consists of plankton images. Moreover, in real plankton recognition applications, a large amount of labeled data for the target dataset is often not available, and the goal is to minimize the required labeling efforts. Therefore, it is more interesting to study the ability of the pretrained models to learn from a small number of labeled images.

In addition to evaluating the models with the full dataset, we experimented using only small subsets of labeled data. We run three different experiments for each model, utilizing only 1\%, 5\%, and 10\% of training data for fine-tuning. To maintain class representation even in the smallest subsets, we ensured a minimum of one sample per class in each experiment. The details of these dataset subsets for both DAPlankton$_{\textrm{LAB}}$ and DAPlankton$_{\textrm{SEA}}$ are presented in \Cref{tab:lab}.

\begin{table}[htb]
    \centering
    \caption{Accuracy (in \%) for full fine-tuning for DAPlankton$_{\textrm{LAB}}$.}
    \resizebox{\linewidth}{!}{%
    \begin{tabular}{lccc}
        Model & IFCB & CS & FC \\
        \midrule[1.25pt]
        ViT-L (imagenet)        & 99.29 ± 0.13 & 98.49 ± 0.23 & 98.18 ± 0.13 \\
        ViT-L (no-daplankton)   & 99.27 ± 0.20 & 98.42 ± 0.30 & 98.35 ± 0.11 \\
        ViT-L (with-daplankton) & 99.32 ± 0.18 & 99.01 ± 0.09 & 98.21 ± 0.19 \\
    \end{tabular}}
    \label{tab:result_full}
\end{table}

\begin{table}[htb]
    \centering
    \caption{Number of labeled samples across DAPlankton$_{\textrm{LAB}}$ and DAPlankton$_{\textrm{SEA}}$ subsets.}
    \resizebox{\linewidth}{!}{%
    \begin{tabular}{c|cc|cc|cc}
        & \multicolumn{2}{c}{1\%} & \multicolumn{2}{|c|}{5\%}  & \multicolumn{2}{c}{10\%}  \\ 
        Dataset & Per Class & Total & Per Class & Total & Per Class & Total \\ 
        \midrule[1.25pt]
        DAPlankton$_{\textrm{LAB}}$ & & & & & &  \\
        IFCB    & 6 -- 9  & 104 & 34 -- 46 & 555& 68 -- 93 & 1,115\\ 
        CS      & 4 -- 8  & 82 & 20 -- 43 & 489 & 41 -- 87 & 888 \\
        FC      & 1 -- 17 & 113 & 9 -- 89 & 598 & 19 -- 178& 1,204\\
        \cmidrule{1-7}
        DAPlankton$_{\textrm{SEA}}$ & & & & & &  \\
        IFCB    & 1 -- 83  & 344 & 1 -- 417 & 1,740 & 3 -- 835 & 3,495\\ 
        CS      & 1 -- 37  & 95  & 1 -- 185 & 430  & 1 -- 370 & 862 \\
    \end{tabular}}
    \label{tab:lab}
\end{table}

Results for DAPlankton$_{\textrm{LAB}}$ are shown in \Cref{fig:ft_lab} for all three imaging instruments. The results show that the MAE models pre-trained with plankton data achieve strong performance even with limited labeled data, clearly outperforming ViT-L pretrained on ImageNet. For example, the model pretrained on all plankton data, ViT-L (with-daplankton), obtained 97\% recognition accuracy on the IFCB subset with only 6-9 labeled training images per class.
ViT-L (with-daplankton) outperforms other pretraining strategies as expected. However, it should be noted that this model has seen the images, albeit without the labels, in the target dataset during pretraining, giving it an advantage over the corresponding models. This corresponds to a scenario where the target dataset is available during the pretraining stage, allowing the model to be exposed to the entire testing distribution during self-supervised pretraining. 

\begin{figure}[htb]
    \centering
    \includegraphics[width=\linewidth]{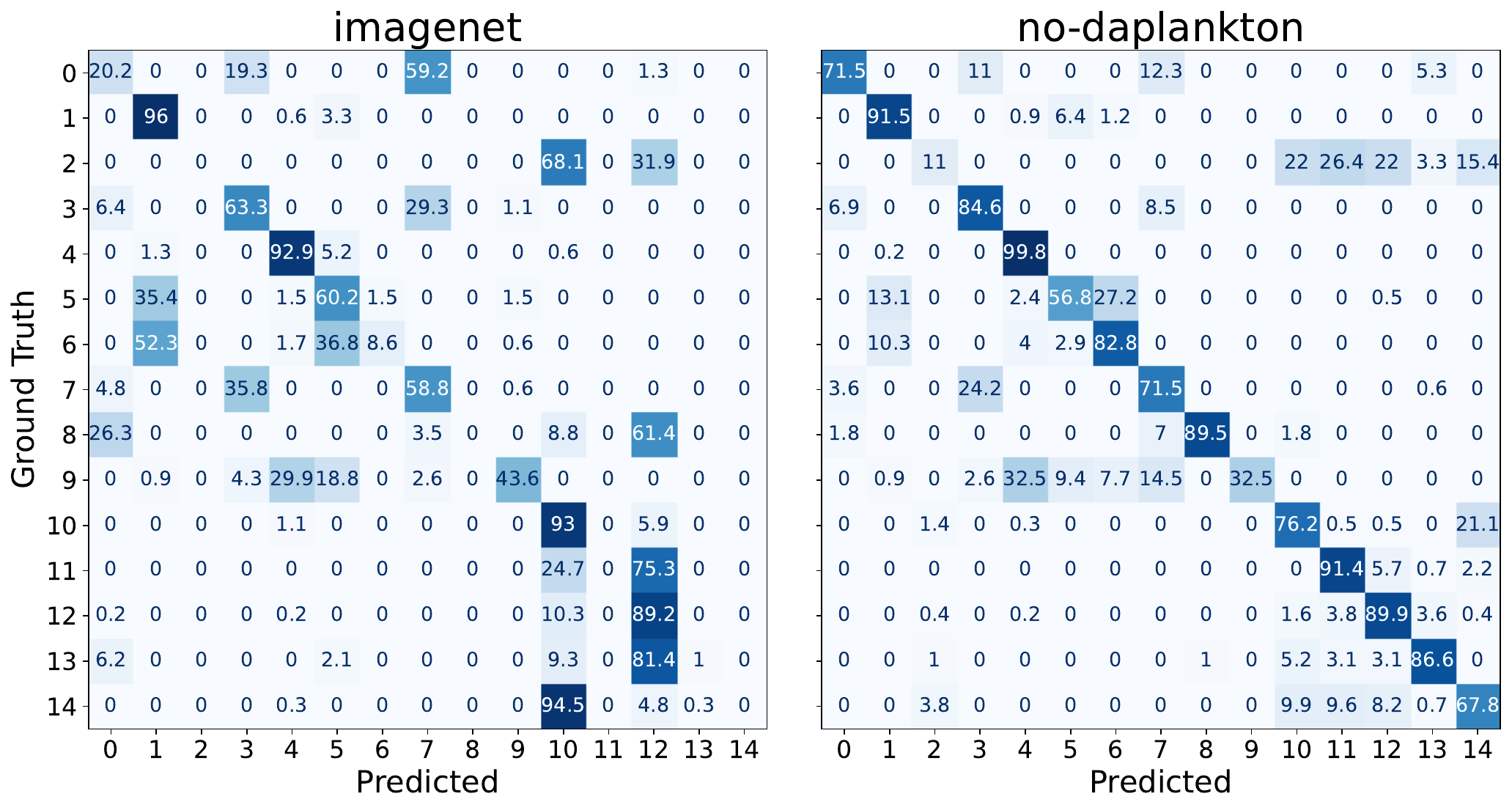}
    \caption{Confusion matrices for ViT-L (ImageNet) and ViT-L (no-daplankton), evaluated on 1\% of labeled FC data from DAPlankton$_\textrm{LAB}$.}
    \label{fig:fc_0.01}
\end{figure}

The model pretrained on all plankton data except DAPlankton, ViT-L (no-daplankton), provides a more realistic scenario where pretraining is done only once, and the model is fine-tuned to target datasets that were not available during pretraining. As can be seen from the results, this model still clearly outperforms the ViT-L model pretrained on ImageNet-1K across all imaging instruments, especially when only 1\% of the training data is used. The confusion matrix with class-specific accuracies is shown in \Cref{fig:fc_0.01}, where only 1\% of the FC data is used. The results indicate that ViT-L (no-daplankton) achieves significantly higher class accuracies and exhibits less confusion overall. In contrast, ViT-L (ImageNet) performs well on only a subset of classes, and it struggles to distinguish between visually similar categories. 

\begin{figure}[htb]
    \centering
    \includegraphics[width=\columnwidth]{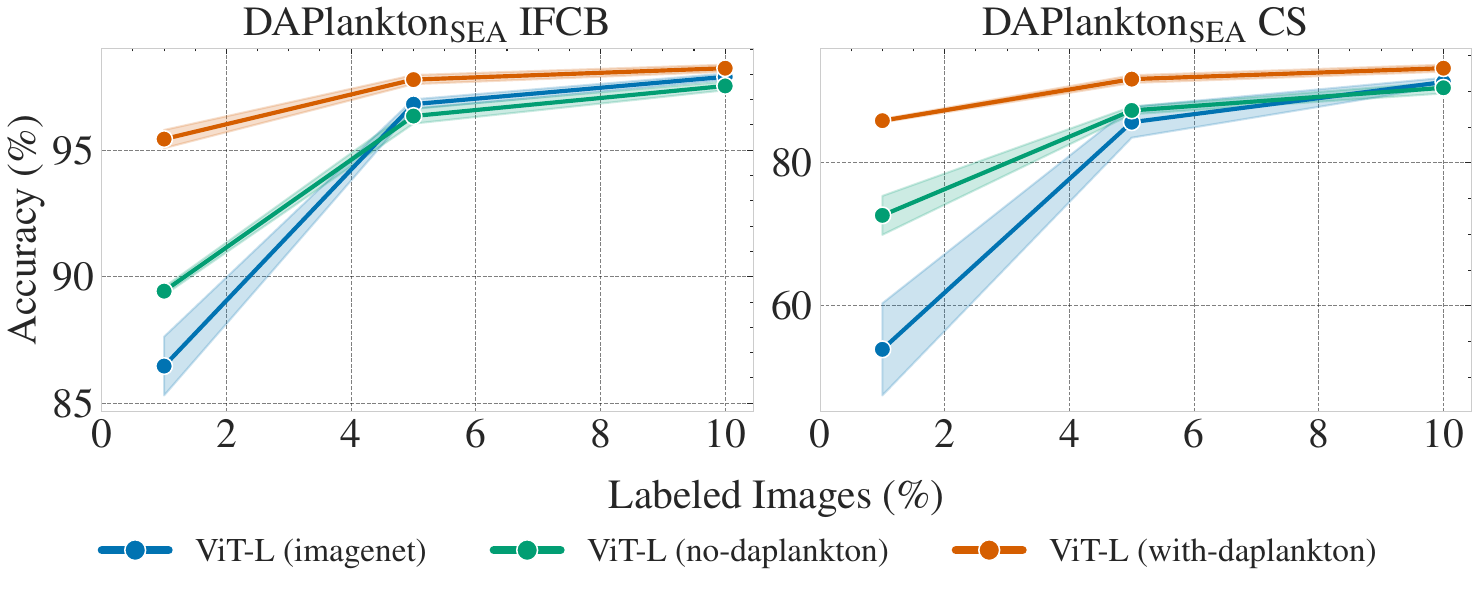}
    \caption{Mean accuracy and standard deviation for DAPlankton$_{\textrm{SEA}}$ across different labeled data subsets.}
    \label{fig:ft_sea}
\end{figure}

Results for fine-tuning with the DAPLankton$_{\textrm{SEA}}$ are shown in \Cref{fig:ft_sea}. The results show that ViT-L (with-daplankton) obtains the best results, followed by ViT-L (no-daplankton) and ViT-L (imagenet). The performance difference is more notable in the CS subset, likely due to its higher class imbalance.  Additionally, ViT-L (imagenet) tends to predict the most common classes and fails to generalize well, resulting in lower accuracy. Overall, the results obtained with DAPlankton$_{\textrm{LAB}}$ and DAPlankton$_{\textrm{SEA}}$ demonstrate the benefit of creating custom pretrained models for plankton recognition, especially if only a small amount of labeled data is available. We have made our pretrained models available for other researchers~\cite{hf-with-daplankton,hf-no-daplankton}.

%% file: sec/5_conclusion.tex
\section{Conclusion}
\label{sec:conclusion}

In this paper, we studied the benefits of self-supervised pretraining for fine-grained plankton recognition. We collected a large and diverse plankton image dataset comprising 443,000 images by combining multiple publicly available datasets captured with different imaging instruments. Due to the varying species compositions and labeling practices, the class labels do not correspond between the individual datasets, making the combined dataset unsuitable for supervised training. Therefore, we applied self-supervised learning using masked autoencoders to obtain general-purpose plankton image encoders.

We evaluated the obtained pretrained encoders by fine-tuning them for plankton recognition on individual datasets with varying amounts of labeled data. When a large amount of labeled training data was used for fine-tuning, the custom pretrained models did not provide a significant benefit over standard ImageNet pretraining. However, when only a small number of labeled training images was used for fine-tuning, the models pretrained on plankton data showed considerably higher recognition accuracy compared to the ImageNet encoder. This highlights the advantages and potential of custom-pretrained models for the fine-grained recognition of plankton images. The model that had access to the unlabeled target data showed the highest recognition accuracy, suggesting that it is beneficial to repeat the pretraining before analyzing new datasets when possible. Nevertheless, the custom model that did not include target data during pretraining also showed very promising results, indicating that self-supervised learning enables the models to learn efficient and general plankton image representations. The results demonstrate that self-supervised learning has the potential to significantly reduce the need for manual labeling of plankton images by experts.

\section{Acknowledgments}
The research was carried out in the FASTVISION and FASTVISION-plus projects funded by the Academy of Finland (Decision numbers 321980, 321991, 339612, and 339355). We wish to acknowledge CSC – IT Center for Science, Finland, for computational resources.

%% file: sec/X_suppl.tex
\clearpage
\setcounter{page}{1}
\onecolumn

\begin{center}
    \Large
    \textbf{\thetitle}\\
    \vspace{0.5em}Supplementary Material \\
    \vspace{1.0em}
\end{center}

\section*{S1. Pretraining dataset class overlap}

The class overlaps between different datasets are shown in Table \ref{tab:overlapping-classes}. In total, there are not many overlapping classes except with DAPlankton and SYKE-Plankton\_IFCB\_2022, which share most of their classes. 

\begin{table}[htb]
\centering
\caption{Overlapping Classes Between Plankton Datasets}
\resizebox{\linewidth}{!}{%
\label{tab:overlapping-classes}
\small
    \begin{tabular}{lll}
    \toprule
    \textbf{Dataset Pair} & \textbf{Overlapping Classes} & \textbf{Count} \\
    \midrule
    DAPlankton~\cite{batrakhanov2024daplankton} \& SYKE-Plankton\_IFCB\_2022~\cite{syke2022} & 
    \begin{tabular}[t]{@{}l@{}}
        aphanizomenon-flosaquae, centrales-sp, chaetoceros-sp, \\
        chaetoceros-sp-single, chlorococcales, chroococcales \\
        ciliata, cryptomonadales, cryptophyceae-teleaulax \\
        cyclotella-choctawhatcheeana, dinophyceae, dinophysis-acuminata \\
        dolichospermum-anabaenopsis, dolichospermum-anabaenopsis-coiled \\
        euglenophyceae, eutreptiella-sp, gymnodiniales,  \\
        gymnodinium-like, heterocapsa-rotundata, heterocapsa-triquetra\\
        heterocyte, katablepharis-remigera, melosira-arctica \\
        mesodinium-rubrum, monoraphidium-contortum, nitzschia-paleacea \\
        nodularia-spumigena, oocystis-sp \\
        peridiniella-catenata (as peridiniella-catenata-single/chain in SYKE) \\
        pseudopedinella-sp, pyramimonas-sp, skeletonema-marinoi \\
        snowella-woronichinia 
    \end{tabular} & 33 \\
    \midrule
    DAPlankton~\cite{batrakhanov2024daplankton}  \& PMID2019~\cite{li2020developing} & chaetoceros (chaetoceros-sp in DA), skeletonema (skeletonema-marinoi in DA)  & 2 \\
    \midrule
    SYKE-Plankton\_IFCB\_2022~\cite{syke2022} \& PMID2019~\cite{li2020developing} & chaetoceros (chaetoceros-sp in SYKE), skeletonema (skeletonema-marinoi in SYKE) & 2 \\
    \midrule
    PMID2019~\cite{li2020developing} \& UDE Diatoms~\cite{Kloster2024} & navicula & 1 \\
    \midrule
    SYKE-Plankton\_ZooScan\_2024~\cite{zooscan2024} \& Kaggle Plankton~\cite{Cowen2015}  & copepod-calanoid, copepod-cyclopoid (multiple cyclopoid in Kaggle) & 2 \\
    \midrule
    SYKE-Plankton\_ZooScan\_2024~\cite{zooscan2024} \& Lake Zooplankton~\cite{kyathanahally2021deep} & daphnia, synchaeta (daphnia-sp, synchaeta-sp in SYKE) & 2 \\
    \midrule
    Lake Zooplankton~\cite{kyathanahally2021deep} \& Kaggle Plankton~\cite{Cowen2015}  & diatom-chain (diatom-chain-string/tube in Kaggle) & 1 \\
    \bottomrule
    \end{tabular}}
\end{table}

\section*{S2. Results}
The fine-tuning results for DAPlankton$_{\textrm{LAB}}$ and DAPlankton$_{\textrm{SEA}}$ are shown in \Cref{tab:results_lab} and \Cref{tab:results_sea}, respectively. For DAPlankton$_{\textrm{SEA}}$, the full fine-tuning results look very similar to DAPlankton$_{\textrm{LAB}}$, and there is no real difference between the different pretraining methods.

\begin{table}[htb]
    \centering
    \caption{Mean accuracy and standard deviation (in \%) for DAPlankton$_{\textrm{LAB}}$ with limited training labels.}
    \resizebox{\linewidth}{!}{%
    \begin{tabular}{l|ccc|ccc|ccc}
        & \multicolumn{3}{c}{IFCB} & \multicolumn{3}{|c|}{CS} & \multicolumn{3}{c}{FC} \\
        Model & 1\% & 5\% & 10\% & 1\% & 5\% & 10\% & 1\% & 5\% & 10\% \\
        \midrule[1.25pt]
        ViT-L (imagenet)      & 78.08 ± 3.54 & 95.23 ± 0.25 & 97.51 ± 0.26 & 62.87 ± 7.45 & 91.63 ± 1.12 & 95.22 ± 0.76 & 52.04 ± 4.39 & 91.72 ± 0.34 & 94.31 ± 0.35 \\
        ViT-L (no-daplankton) & 83.83 ± 2.03 & 95.68 ± 0.35 & 97.41 ± 0.23 & 69.90 ± 4.73 & 90.30 ± 0.80 & 93.57 ± 0.47 & 79.58 ± 2.58 & 92.87 ± 0.33 & 95.00 ± 0.48 \\
        ViT-L (with-daplankton)      & 96.92 ± 0.37 & 98.29 ± 0.30 & 98.75 ± 0.24 & 91.76 ± 0.89 & 96.26 ± 0.52 & 96.97 ± 0.33 & 83.10 ± 1.44 & 93.86 ± 0.65 & 95.72 ± 0.64 \\
    \end{tabular}}
    \label{tab:results_lab}
\end{table}

\begin{table}[htb]
    \centering
    \caption{Mean accuracy and standard deviation (in \%) for DAPlankton$_{\textrm{SEA}}$ with limited training labels.}
    \resizebox{\linewidth}{!}{%
    \begin{tabular}{l|cccc|cccc}
        & \multicolumn{4}{c}{IFCB} & \multicolumn{4}{|c}{CS}\\
        Model & 1\% & 5\% & 10\% & Full & 1\% & 5\% & 10\% & Full \\
        \midrule[1.25pt]
        ViT-L (imagenet)      & 86.47 ± 1.16 & 96.82 ± 0.19 & 97.90 ± 0.13 & 98.91 ± 0.04 & 53.92 ± 6.44 & 85.68 ± 2.17 & 91.29 ± 0.59 & 95.80 ± 0.44 \\
        ViT-L (no-daplankton) & 89.43 ± 0.16 & 96.35 ± 0.30 & 97.54 ± 0.17 & 98.84 ± 0.10 & 72.65 ± 2.72 & 87.32 ± 0.54 & 90.49 ± 0.79 & 95.32 ± 0.06  \\
        ViT-L (with-daplankton)      & 95.44 ± 0.36 & 97.79 ± 0.18 & 98.24 ± 0.14 & 98.85 ± 0.06 & 85.89 ± 0.24 & 91.70 ± 0.54 & 93.23 ± 0.49 & 95.57 ± 0.24  \\
    \end{tabular}}
    \label{tab:results_sea}
\end{table}

%% file: main.bbl
\begin{thebibliography}{71}
\providecommand{\natexlab}[1]{#1}
\providecommand{\url}[1]{\texttt{#1}}
\expandafter\ifx\csname urlstyle\endcsname\relax
  \providecommand{\doi}[1]{doi: #1}\else
  \providecommand{\doi}{doi: \begingroup \urlstyle{rm}\Url}\fi

\bibitem[Badreldeen Bdawy~Mohamed et~al.(2022)Badreldeen Bdawy~Mohamed, Eerola, Kraft, Lensu, and K{\"a}lvi{\"a}inen]{badreldeen2022open}
Ola Badreldeen Bdawy~Mohamed, Tuomas Eerola, Kaisa Kraft, Lasse Lensu, and Heikki K{\"a}lvi{\"a}inen.
\newblock Open-set plankton recognition using similarity learning.
\newblock In \emph{International Symposium on Visual Computing}, pages 174--183, 2022.

\bibitem[Banse(1995)]{Banse1995}
Karl Banse.
\newblock Zooplankton: Pivotal role in the control of ocean production: I. biomass and production.
\newblock \emph{ICES Journal of Marine Science}, 52:\penalty0 265--277, 1995.

\bibitem[Bao et~al.(2021)Bao, Dong, Piao, and Wei]{bao2021beit}
Hangbo Bao, Li Dong, Songhao Piao, and Furu Wei.
\newblock {BEIT: BERT} pre-training of image transformers.
\newblock https://arxiv.org/abs/2106.08254, 2021.

\bibitem[Batrakhanov et~al.(2024)Batrakhanov, Eerola, Kraft, Haraguchi, Lensu, Suikkanen, Camarena-G{\'o}mez, Sepp{\"a}l{\"a}, and K{\"a}lvi{\"a}inen]{batrakhanov2024daplankton}
Daniel Batrakhanov, Tuomas Eerola, Kaisa Kraft, Lumi Haraguchi, Lasse Lensu, Sanna Suikkanen, Mar{\'\i}a~Teresa Camarena-G{\'o}mez, Jukka Sepp{\"a}l{\"a}, and Heikki K{\"a}lvi{\"a}inen.
\newblock {DAPlankton}: Benchmark dataset for multi-instrument plankton recognition via fine-grained domain adaptation.
\newblock In \emph{ICIP}, pages 158--164, 2024.

\bibitem[Beaugrand et~al.(2003)Beaugrand, Brander, Alistair~Lindley, Souissi, and Reid]{beaugrand2003plankton}
Gr{\'e}gory Beaugrand, Keith~M Brander, J Alistair~Lindley, Sami Souissi, and Philip~C Reid.
\newblock Plankton effect on cod recruitment in the north sea.
\newblock \emph{Nature}, 426:\penalty0 661--664, 2003.

\bibitem[Bure{\v{s}} et~al.(2021)Bure{\v{s}}, Eerola, Lensu, K{\"a}lvi{\"a}inen, and Zem{\v{c}}{\'\i}k]{burevs2021plankton}
Jaroslav Bure{\v{s}}, Tuomas Eerola, Lasse Lensu, Heikki K{\"a}lvi{\"a}inen, and Pavel Zem{\v{c}}{\'\i}k.
\newblock Plankton recognition in images with varying size.
\newblock In \emph{ICPR Workshops and Challenges}, pages 110--120, 2021.

\bibitem[Callejas et~al.(2025)Callejas, Lira, Berry, Mart{\'i}, and Sanchez-Pi]{callejas2025}
Sof{\'i}a Callejas, Hernan Lira, Andrew Berry, Luis Mart{\'i}, and Nayat Sanchez-Pi.
\newblock No plankton left behind: Preliminary results on massive plankton image recognition.
\newblock In \emph{High Performance Computing}, pages 170--185, 2025.

\bibitem[Caron et~al.(2021)Caron, Touvron, Misra, J{\'e}gou, Mairal, Bojanowski, and Joulin]{caron2021dino}
Mathilde Caron, Hugo Touvron, Ishan Misra, Herv{\'e} J{\'e}gou, Julien Mairal, Piotr Bojanowski, and Armand Joulin.
\newblock Emerging properties in self-supervised vision transformers.
\newblock In \emph{ICCV}, pages 9650--9660, 2021.

\bibitem[Chen et~al.(2020{\natexlab{a}})Chen, Kornblith, Norouzi, and Hinton]{chen2020simple}
Ting Chen, Simon Kornblith, Mohammad Norouzi, and Geoffrey Hinton.
\newblock A simple framework for contrastive learning of visual representations.
\newblock In \emph{International Conference on Machine Learning}, pages 1597--1607, 2020{\natexlab{a}}.

\bibitem[Chen et~al.(2020{\natexlab{b}})Chen, Kornblith, Swersky, Norouzi, and Hinton]{chen2020big}
Ting Chen, Simon Kornblith, Kevin Swersky, Mohammad Norouzi, and Geoffrey~E Hinton.
\newblock Big self-supervised models are strong semi-supervised learners.
\newblock \emph{NeurIPS}, 33:\penalty0 22243--22255, 2020{\natexlab{b}}.

\bibitem[Cowen et~al.(2015)Cowen, Sponaugle, Robinson, Luo, and Guigand]{Cowen2015}
Robert~K. Cowen, Su Sponaugle, Kelly~L. Robinson, Jessica Luo, and Cedric Guigand.
\newblock Planktonset 1.0: Plankton imagery data collected from f.g. walton smith in straits of florida from 2014-06-03 to 2014-06-06 and used in the 2015 national data science bowl (ncei accession 0127422).
\newblock \url{https://doi.org/10.7289/v5d21vjd}, 2015.

\bibitem[Devlin et~al.(2019)Devlin, Chang, Lee, and Toutanova]{devlin2019bert}
Jacob Devlin, Ming-Wei Chang, Kenton Lee, and Kristina Toutanova.
\newblock {BERT}: Pre-training of deep bidirectional transformers for language understanding.
\newblock In \emph{North American Chapter of the Association for Computational Linguistics}, 2019.

\bibitem[Duan et~al.(2024)Duan, Zhao, Qi, Zhou, Wang, and Shi]{duan2024roll}
Yue Duan, Zhen Zhao, Lei Qi, Luping Zhou, Lei Wang, and Yinghuan Shi.
\newblock Roll with the punches: expansion and shrinkage of soft label selection for semi-supervised fine-grained learning.
\newblock In \emph{AAAI Conference on Artificial Intelligence}, pages 11829--11837, 2024.

\bibitem[Dubelaar et~al.(1999)Dubelaar, Gerritzen, Beeker, Jonker, and Tangen]{dubelaar1999design}
George~BJ Dubelaar, Peter~L Gerritzen, Arnout~ER Beeker, Richard~R Jonker, and Karl Tangen.
\newblock Design and first results of cytobuoy: A wireless flow cytometer for in situ analysis of marine and fresh waters.
\newblock \emph{Cytometry}, 37:\penalty0 247--254, 1999.

\bibitem[Eerola et~al.(2024)Eerola, Batrakhanov, Barazandeh, Kraft, Haraguchi, Lensu, Suikkanen, Sepp{\"a}l{\"a}, Tamminen, and K{\"a}lvi{\"a}inen]{eerola2024survey}
Tuomas Eerola, Daniel Batrakhanov, Nastaran~Vatankhah Barazandeh, Kaisa Kraft, Lumi Haraguchi, Lasse Lensu, Sanna Suikkanen, Jukka Sepp{\"a}l{\"a}, Timo Tamminen, and Heikki K{\"a}lvi{\"a}inen.
\newblock Survey of automatic plankton image recognition: challenges, existing solutions and future perspectives.
\newblock \emph{Artificial Intelligence Review}, 57:\penalty0 114, 2024.

\bibitem[Elineau et~al.(2024)Elineau, Desnos, Jalabert, Olivier, Romagnan, Costa~Brandao, Lombard, Llopis, Courboulès, Caray-Counil, Serranito, Irisson, Picheral, Gorsky, and Stemmann]{zooscannet}
Amanda Elineau, Corinne Desnos, Laetitia Jalabert, Marion Olivier, Jean-Baptiste Romagnan, Manoela Costa~Brandao, Fabien Lombard, Natalia Llopis, Justine Courboulès, Louis Caray-Counil, Bruno Serranito, Jean-Olivier Irisson, Marc Picheral, Gaby Gorsky, and Lars Stemmann.
\newblock Zooscannet: plankton images captured with the zooscan.
\newblock \url{https://doi.org/10.17882/55741}, 2024.

\bibitem[Falkowski(1994)]{falkowski1994role}
Paul~G Falkowski.
\newblock The role of phytoplankton photosynthesis in global biogeochemical cycles.
\newblock \emph{Photosynthesis research}, 39:\penalty0 235--258, 1994.

\bibitem[Field et~al.(1998)Field, Behrenfeld, Randerson, and Falkowski]{field1998primary}
Christopher~B Field, Michael~J Behrenfeld, James~T Randerson, and Paul Falkowski.
\newblock Primary production of the biosphere: integrating terrestrial and oceanic components.
\newblock \emph{Science}, 281:\penalty0 237--240, 1998.

\bibitem[Goodfellow et~al.(2014)Goodfellow, Pouget-Abadie, Mirza, Xu, Warde-Farley, Ozair, Courville, and Bengio]{goodfellow2014generative}
Ian Goodfellow, Jean Pouget-Abadie, Mehdi Mirza, Bing Xu, David Warde-Farley, Sherjil Ozair, Aaron Courville, and Yoshua Bengio.
\newblock Generative adversarial nets.
\newblock \emph{NeurIPS}, 27, 2014.

\bibitem[Goyal et~al.(2017)Goyal, Doll{\'a}r, Girshick, Noordhuis, Wesolowski, Kyrola, Tulloch, Jia, and He]{goyal2017accurate}
Priya Goyal, Piotr Doll{\'a}r, Ross Girshick, Pieter Noordhuis, Lukasz Wesolowski, Aapo Kyrola, Andrew Tulloch, Yangqing Jia, and Kaiming He.
\newblock Accurate, large minibatch sgd: Training imagenet in 1 hour.
\newblock \url{https://arxiv.org/abs/1706.02677}, 2017.

\bibitem[Grill et~al.(2020)Grill, Strub, Altch{\'e}, Tallec, Richemond, Buchatskaya, Doersch, Avila~Pires, Guo, Gheshlaghi~Azar, et~al.]{grill2020bootstrap}
Jean-Bastien Grill, Florian Strub, Florent Altch{\'e}, Corentin Tallec, Pierre Richemond, Elena Buchatskaya, Carl Doersch, Bernardo Avila~Pires, Zhaohan Guo, Mohammad Gheshlaghi~Azar, et~al.
\newblock Bootstrap your own latent-a new approach to self-supervised learning.
\newblock \emph{NeurIPS}, 33:\penalty0 21271--21284, 2020.

\bibitem[Hays et~al.(2005)Hays, Richardson, and Robinson]{hays2005}
Graeme Hays, Anthony Richardson, and Carol Robinson.
\newblock Climate change and marine plankton.
\newblock \emph{Trends in Ecology \& Evolution}, 20:\penalty0 337--344, 2005.

\bibitem[He et~al.(2020)He, Fan, Wu, Xie, and Girshick]{he2020momentum}
Kaiming He, Haoqi Fan, Yuxin Wu, Saining Xie, and Ross Girshick.
\newblock Momentum contrast for unsupervised visual representation learning.
\newblock In \emph{CVPR}, pages 9729--9738, 2020.

\bibitem[He et~al.(2022)He, Chen, Xie, Li, Doll{\'a}r, and Girshick]{he2022masked}
Kaiming He, Xinlei Chen, Saining Xie, Yanghao Li, Piotr Doll{\'a}r, and Ross Girshick.
\newblock Masked autoencoders are scalable vision learners.
\newblock In \emph{CVPR}, pages 16000--16009, 2022.

\bibitem[Hendrycks and Gimpel(2016)]{GELU2016}
Dan Hendrycks and Kevin Gimpel.
\newblock {Gaussian error linear units (GELUs)}.
\newblock \url{https://arxiv.org/abs/1606.08415}, 2016.

\bibitem[Huang et~al.(2016)Huang, Sun, Liu, Sedra, and Weinberger]{huang2016deep}
Gao Huang, Yu Sun, Zhuang Liu, Daniel Sedra, and Kilian~Q Weinberger.
\newblock Deep networks with stochastic depth.
\newblock In \emph{ECCV}, pages 646--661, 2016.

\bibitem[Jaiswal et~al.(2020)Jaiswal, Babu, Zadeh, Banerjee, and Makedon]{jaiswal2020survey}
Ashish Jaiswal, Ashwin~Ramesh Babu, Mohammad~Zaki Zadeh, Debapriya Banerjee, and Fillia Makedon.
\newblock A survey on contrastive self-supervised learning.
\newblock \emph{Technologies}, 9:\penalty0 2, 2020.

\bibitem[Kareinen et~al.(2024{\natexlab{a}})Kareinen, Skytt{\"a}, Eerola, Kraft, Lensu, Suikkanen, Lehtiniemi, and K{\"a}lvi{\"a}inen]{kareinen2024open}
Joona Kareinen, Annaliina Skytt{\"a}, Tuomas Eerola, Kaisa Kraft, Lasse Lensu, Sanna Suikkanen, Maiju Lehtiniemi, and Heikki K{\"a}lvi{\"a}inen.
\newblock Open-set plankton recognition.
\newblock In \emph{ECCV Workshops}, 2024{\natexlab{a}}.

\bibitem[Kareinen et~al.(2024{\natexlab{b}})Kareinen, Skyttä, abd Kaisa~Kraft, Lensu, Suikkanen, Lehtiniemi, and Kälviäinen]{zooscan2024}
Joona Kareinen, Annaliina Skyttä, Tuomas~Eerola abd Kaisa~Kraft, Lasse Lensu, Sanna Suikkanen, Maiju Lehtiniemi, and Heikki Kälviäinen.
\newblock {SYKE-plankton\_ZooScan\_2024}.
\newblock \url{https://doi.org/10.23729/fa115087-2698-4aa5-aedd-11e260b9694d}, 2024{\natexlab{b}}.

\bibitem[Kareinen et~al.(2025{\natexlab{a}})Kareinen, Eerola, Kraft, Lensu, Suikkanen, and Kälviäinen]{hf-no-daplankton}
Joona Kareinen, Tuomas Eerola, Kaisa Kraft, Lasse Lensu, Sanna Suikkanen, and Heikki Kälviäinen.
\newblock no-daplankton.
\newblock \url{https://huggingface.co/Jookare/no_daplankton_vit_large_patch16_224.mae}, 2025{\natexlab{a}}.
\newblock [Online; accessed April, 11, 2025].

\bibitem[Kareinen et~al.(2025{\natexlab{b}})Kareinen, Eerola, Kraft, Lensu, Suikkanen, and Kälviäinen]{hf-with-daplankton}
Joona Kareinen, Tuomas Eerola, Kaisa Kraft, Lasse Lensu, Sanna Suikkanen, and Heikki Kälviäinen.
\newblock with-daplankton.
\newblock \url{https://huggingface.co/Jookare/plankton_vit_large_patch16_224.mae}, 2025{\natexlab{b}}.
\newblock [Online; accessed April, 11, 2025].

\bibitem[Kim et~al.(2023)Kim, Bae, and Yun]{kim2023coreset}
Sungnyun Kim, Sangmin Bae, and Se-Young Yun.
\newblock Coreset sampling from open-set for fine-grained self-supervised learning.
\newblock In \emph{CVPR}, pages 7537--7547, 2023.

\bibitem[Kloster et~al.(2024)Kloster, Burfeid-Castellanos, Dani, Mayombo, Beszteri, and Vidaković]{Kloster2024}
Michael Kloster, Andrea Burfeid-Castellanos, Mimoza Dani, Ntambwe Albert~Serge Mayombo, Bank Beszteri, and Danijela Vidaković.
\newblock {UDE Diatoms in the Wild 2024}, 2024.

\bibitem[Kolesnikov et~al.(2021)Kolesnikov, Dosovitskiy, Weissenborn, Heigold, Uszkoreit, Beyer, Minderer, Dehghani, Houlsby, Gelly, Unterthiner, and Zhai]{Kolesnikov2021vit}
Alexander Kolesnikov, Alexey Dosovitskiy, Dirk Weissenborn, Georg Heigold, Jakob Uszkoreit, Lucas Beyer, Matthias Minderer, Mostafa Dehghani, Neil Houlsby, Sylvain Gelly, Thomas Unterthiner, and Xiaohua Zhai.
\newblock An image is worth 16x16 words: Transformers for image recognition at scale.
\newblock In \emph{ICLR}, 2021.

\bibitem[Kraft et~al.(2022{\natexlab{a}})Kraft, Velhonoja, Eerola, Suikkanen, Tamminen, Haraguchi, Yl{\"o}stalo, Kielosto, Johansson, Lensu, et~al.]{kraft2022towards}
Kaisa Kraft, Otso Velhonoja, Tuomas Eerola, Sanna Suikkanen, Timo Tamminen, Lumi Haraguchi, Pasi Yl{\"o}stalo, Sami Kielosto, Milla Johansson, Lasse Lensu, et~al.
\newblock Towards operational phytoplankton recognition with automated high-throughput imaging, near-real-time data processing, and convolutional neural networks.
\newblock \emph{Frontiers in Marine Science}, 9, 2022{\natexlab{a}}.

\bibitem[Kraft et~al.(2022{\natexlab{b}})Kraft, Velhonoja, Seppälä, Hällfors, Suikkanen, Ylöstalo, Anglès, Kielosto, Kuosa, Lehtinen, Oja, and Tamminen]{syke2022}
Kaisa Kraft, Otso Velhonoja, Jukka Seppälä, Heidi Hällfors, Sanna Suikkanen, Pasi Ylöstalo, Sílvia Anglès, Sami Kielosto, Harri Kuosa, Sirpa Lehtinen, Johanna Oja, and Timo Tamminen.
\newblock {SYKE-plankton\_IFCB\_2022}.
\newblock \url{https://b2share.eudat.eu/records/abf913e5a6ad47e6baa273ae0ed6617a}, 2022{\natexlab{b}}.

\bibitem[Kyathanahally et~al.(2021)Kyathanahally, Hardeman, Merz, Bulas, Reyes, Isles, Pomati, and Baity-Jesi]{kyathanahally2021deep}
Sreenath~P Kyathanahally, Thomas Hardeman, Ewa Merz, Thea Bulas, Marta Reyes, Peter Isles, Francesco Pomati, and Marco Baity-Jesi.
\newblock Deep learning classification of lake zooplankton.
\newblock \emph{Frontiers in Microbiology}, 12:\penalty0 746297, 2021.

\bibitem[Kyathanahally et~al.(2022)Kyathanahally, Hardeman, Reyes, Merz, Bulas, Brun, Pomati, and Baity-Jesi]{kyathanahally2022ensembles}
Sreenath~P Kyathanahally, Thomas Hardeman, Marta Reyes, Ewa Merz, Thea Bulas, Philipp Brun, Francesco Pomati, and Marco Baity-Jesi.
\newblock Ensembles of data-efficient vision transformers as a new paradigm for automated classification in ecology.
\newblock \emph{Scientific Reports}, 12:\penalty0 18590, 2022.

\bibitem[Li et~al.(2020)Li, Sun, Dong, Song, Zhang, Liu, Zhang, and Han]{li2020developing}
Qiong Li, Xin Sun, Junyu Dong, Shuqun Song, Tongtong Zhang, Dan Liu, Han Zhang, and Shuai Han.
\newblock Developing a microscopic image dataset in support of intelligent phytoplankton detection using deep learning.
\newblock \emph{ICES Journal of Marine Science}, 77:\penalty0 1427--1439, 2020.

\bibitem[Lombard et~al.(2019)Lombard, Boss, Waite, Vogt, Uitz, Stemmann, Sosik, Schulz, Romagnan, Picheral, Pearlman, Ohman, Niehoff, Möller, Miloslavich, Lara-Lpez, Kudela, Lopes, Kiko, Karp-Boss, Jaffe, Iversen, Irisson, Fennel, Hauss, Guidi, Gorsky, Giering, Gaube, Gallager, Dubelaar, Cowen, Carlotti, Briseño-Avena, Berline, Benoit-Bird, Bax, Batten, Ayata, Artigas, and Appeltans]{global2019}
Fabien Lombard, Emmanuel Boss, Anya~M. Waite, Meike Vogt, Julia Uitz, Lars Stemmann, Heidi~M. Sosik, Jan Schulz, Jean-Baptiste Romagnan, Marc Picheral, Jay Pearlman, Mark~D. Ohman, Barbara Niehoff, Klas~O. Möller, Patricia Miloslavich, Ana Lara-Lpez, Raphael Kudela, Rubens~M. Lopes, Rainer Kiko, Lee Karp-Boss, Jules~S. Jaffe, Morten~H. Iversen, Jean-Olivier Irisson, Katja Fennel, Helena Hauss, Lionel Guidi, Gaby Gorsky, Sarah L.~C. Giering, Peter Gaube, Scott Gallager, George Dubelaar, Robert~K. Cowen, François Carlotti, Christian Briseño-Avena, Léo Berline, Kelly Benoit-Bird, Nicholas Bax, Sonia Batten, Sakina~Dorothée Ayata, Luis~Felipe Artigas, and Ward Appeltans.
\newblock Globally consistent quantitative observations of planktonic ecosystems.
\newblock \emph{Frontiers in Marine Science}, 6, 2019.

\bibitem[Loshchilov and Hutter(2019)]{loshchilov2018decoupled}
Ilya Loshchilov and Frank Hutter.
\newblock Decoupled weight decay regularization.
\newblock In \emph{ICLR}, 2019.

\bibitem[Lumini and Nanni(2019{\natexlab{a}})]{lumini2019}
Alessandra Lumini and Loris Nanni.
\newblock Deep learning and transfer learning features for plankton classification.
\newblock \emph{Ecological Informatics}, 51:\penalty0 33--43, 2019{\natexlab{a}}.

\bibitem[Lumini and Nanni(2019{\natexlab{b}})]{lumini2019deep}
Alessandra Lumini and Loris Nanni.
\newblock Deep learning and transfer learning features for plankton classification.
\newblock \emph{Ecological Informatics}, 51:\penalty0 33--43, 2019{\natexlab{b}}.

\bibitem[Maracani et~al.(2023)Maracani, Pastore, Natale, Rosasco, and Odone]{maracani2023domain}
Andrea Maracani, Vito~Paolo Pastore, Lorenzo Natale, Lorenzo Rosasco, and Francesca Odone.
\newblock In-domain versus out-of-domain transfer learning in plankton image classification.
\newblock \emph{Scientific Reports}, 13:\penalty0 10443, 2023.

\bibitem[Oldenburg et~al.(2023)Oldenburg, Kronberg, Niehoff, Ebenh{\"o}h, and Popa]{oldenburg2023deeploki}
Ellen Oldenburg, Raphael~M Kronberg, Barbara Niehoff, Oliver Ebenh{\"o}h, and Ovidiu Popa.
\newblock {DeepLOKI-a deep learning based approach to identify zooplankton taxa on high-resolution images from the optical plankton recorder LOKI}.
\newblock \emph{Frontiers in Marine Science}, 10:\penalty0 1280510, 2023.

\bibitem[Olson and Sosik(2007)]{olson2007submersible}
Robert~J Olson and Heidi~M Sosik.
\newblock A submersible imaging-in-flow instrument to analyze nano-and microplankton: Imaging flowcytobot.
\newblock \emph{Limnology and Oceanography: Methods}, 5:\penalty0 195--203, 2007.

\bibitem[Oquab et~al.(2024)Oquab, Darcet, Moutakanni, Vo, Szafraniec, Khalidov, Fernandez, Haziza, Massa, El-Nouby, Assran, Ballas, Galuba, Howes, Huang, Li, Misra, Rabbat, Sharma, Synnaeve, Xu, Jegou, Mairal, Labatut, Joulin, and Bojanowski]{dinov2}
Maxime Oquab, Timothée Darcet, Théo Moutakanni, Huy Vo, Marc Szafraniec, Vasil Khalidov, Pierre Fernandez, Daniel Haziza, Francisco Massa, Alaaeldin El-Nouby, Mahmoud Assran, Nicolas Ballas, Wojciech Galuba, Russell Howes, Po-Yao Huang, Shang-Wen Li, Ishan Misra, Michael Rabbat, Vasu Sharma, Gabriel Synnaeve, Hu Xu, Hervé Jegou, Julien Mairal, Patrick Labatut, Armand Joulin, and Piotr Bojanowski.
\newblock Dinov2: Learning robust visual features without supervision.
\newblock \url{https://arxiv.org/abs/2304.07193}, 2024.

\bibitem[Orenstein and Beijbom(2017)]{orenstein2017transfer}
Eric~C Orenstein and Oscar Beijbom.
\newblock Transfer learning and deep feature extraction for planktonic image data sets.
\newblock In \emph{IEEE Winter Conference on Applications of Computer Vision}, pages 1082--1088, 2017.

\bibitem[Orenstein et~al.(2015)Orenstein, Beijbom, Peacock, and Sosik]{WHOIplankton}
Eric~C. Orenstein, Oscar Beijbom, Emily~E. Peacock, and Heidi~M. Sosik.
\newblock Whoi-plankton- a large scale fine grained visual recognition benchmark dataset for plankton classification.
\newblock \url{https://arxiv.org/abs/1510.00745}, 2015.

\bibitem[Pastore et~al.(2023)Pastore, Ciranni, Bianco, Fung, Murino, and Odone]{pastore2023efficient}
Vito~Paolo Pastore, Massimiliano Ciranni, Simone Bianco, Jennifer~Carol Fung, Vittorio Murino, and Francesca Odone.
\newblock Efficient unsupervised learning of biological images with compressed deep features.
\newblock \emph{Image and Vision Computing}, 137:\penalty0 104764, 2023.

\bibitem[Peng et~al.(2022)Peng, Dong, Bao, Ye, and Wei]{beitv2}
Zhiliang Peng, Li Dong, Hangbo Bao, Qixiang Ye, and Furu Wei.
\newblock Beit v2: Masked image modeling with vector-quantized visual tokenizers.
\newblock \url{https://arxiv.org/abs/2208.06366}, 2022.

\bibitem[Pu et~al.(2021)Pu, Feng, Wang, Yang, and Li]{pu2021anomaly}
Yuchun Pu, Zhenghui Feng, Zhonglei Wang, Zhenyu Yang, and Jianping Li.
\newblock Anomaly detection for in situ marine plankton images.
\newblock In \emph{ICCV}, pages 3661--3671, 2021.

\bibitem[Salvesen et~al.(2020)Salvesen, Saad, and Stahl]{salvesen2020robust}
Eivind Salvesen, Aya Saad, and Annette Stahl.
\newblock Robust methods of unsupervised clustering to discover new planktonic species in-situ.
\newblock In \emph{Global Oceans 2020: Singapore--US Gulf Coast}, pages 1--9, 2020.

\bibitem[Salvesen et~al.(2022)Salvesen, Saad, and Stahl]{salvesen2022robust}
Eivind Salvesen, Aya Saad, and Annette Stahl.
\newblock Robust deep unsupervised learning framework to discover unseen plankton species.
\newblock In \emph{International Conference on Machine Vision}, pages 241--250, 2022.

\bibitem[Schanz et~al.(2023)Schanz, M{\"o}ller, R{\"u}hl, and Greenberg]{schanz2023robust}
Tobias Schanz, Klas~Ove M{\"o}ller, Saskia R{\"u}hl, and David~S Greenberg.
\newblock Robust detection of marine life with label-free image feature learning and probability calibration.
\newblock \emph{Machine Learning: Science and Technology}, 4:\penalty0 035007, 2023.

\bibitem[Schmarje et~al.(2021)Schmarje, Br{\"u}nger, Santarossa, Schr{\"o}der, Kiko, and Koch]{schmarje2021fuzzy}
Lars Schmarje, Johannes Br{\"u}nger, Monty Santarossa, Simon-Martin Schr{\"o}der, Rainer Kiko, and Reinhard Koch.
\newblock Fuzzy overclustering: Semi-supervised classification of fuzzy labels with overclustering and inverse cross-entropy.
\newblock \emph{Sensors}, 21:\penalty0 6661, 2021.

\bibitem[Schmidhuber(2015)]{schmidhuber2015deep}
J{\"u}rgen Schmidhuber.
\newblock Deep learning in neural networks: An overview.
\newblock \emph{Neural Networks}, 61:\penalty0 85--117, 2015.

\bibitem[Schr{\"o}der and Kiko(2022)]{schroder2022assessing}
Simon-Martin Schr{\"o}der and Rainer Kiko.
\newblock Assessing representation learning and clustering algorithms for computer-assisted image annotation—simulating and benchmarking {MorphoCluster}.
\newblock \emph{Sensors}, 22:\penalty0 2775, 2022.

\bibitem[Schr{\"o}der et~al.(2020)Schr{\"o}der, Kiko, and Koch]{schroder2020morphocluster}
Simon-Martin Schr{\"o}der, Rainer Kiko, and Reinhard Koch.
\newblock {MorphoCluster}: efficient annotation of plankton images by clustering.
\newblock \emph{Sensors}, 20:\penalty0 3060, 2020.

\bibitem[Sieracki et~al.(1998)Sieracki, Sieracki, and Yentsch]{sieracki1998imaging}
Christian~K Sieracki, Michael~E Sieracki, and Charles~S Yentsch.
\newblock An imaging-in-flow system for automated analysis of marine microplankton.
\newblock \emph{Marine Ecology Progress Series}, 168:\penalty0 285--296, 1998.

\bibitem[Simon et~al.(2009)Simon, Cras, Foulon, and Lem{\'e}e]{simon2009diversity}
Nathalie Simon, Anne-Lise Cras, Elodie Foulon, and Rodolphe Lem{\'e}e.
\newblock Diversity and evolution of marine phytoplankton.
\newblock \emph{Comptes Rendus Biologies}, 332:\penalty0 159--170, 2009.

\bibitem[Singh et~al.(2023)Singh, Duval, Alwala, Fan, Aggarwal, Adcock, Joulin, Doll{\'a}r, Feichtenhofer, Girshick, et~al.]{singh2023effectiveness}
Mannat Singh, Quentin Duval, Kalyan~Vasudev Alwala, Haoqi Fan, Vaibhav Aggarwal, Aaron Adcock, Armand Joulin, Piotr Doll{\'a}r, Christoph Feichtenhofer, Ross Girshick, et~al.
\newblock The effectiveness of mae pre-pretraining for billion-scale pretraining.
\newblock In \emph{ICCV}, pages 5484--5494, 2023.

\bibitem[Stevens et~al.(2024)Stevens, Wu, Thompson, Campolongo, Song, Carlyn, Dong, Dahdul, Stewart, Berger-Wolf, et~al.]{stevens2024bioclip}
Samuel Stevens, Jiaman Wu, Matthew~J Thompson, Elizabeth~G Campolongo, Chan~Hee Song, David~Edward Carlyn, Li Dong, Wasila~M Dahdul, Charles Stewart, Tanya Berger-Wolf, et~al.
\newblock {BioCLIP: A Vision Foundation Model for the Tree of Life }.
\newblock In \emph{CVPR}, pages 19412--19424, 2024.

\bibitem[Su et~al.(2021)Su, Cheng, and Maji]{su2021realistic}
Jong-Chyi Su, Zezhou Cheng, and Subhransu Maji.
\newblock A realistic evaluation of semi-supervised learning for fine-grained classification.
\newblock In \emph{CVPR}, pages 12966--12975, 2021.

\bibitem[Szegedy et~al.(2016)Szegedy, Vanhoucke, Ioffe, Shlens, and Wojna]{szegedy2016rethinking}
Christian Szegedy, Vincent Vanhoucke, Sergey Ioffe, Jon Shlens, and Zbigniew Wojna.
\newblock Rethinking the inception architecture for computer vision.
\newblock In \emph{CVPR}, pages 2818--2826, 2016.

\bibitem[Touvron et~al.(2021)Touvron, Cord, Douze, Massa, Sablayrolles, and J{\'e}gou]{touvron2021deit}
Hugo Touvron, Matthieu Cord, Matthijs Douze, Francisco Massa, Alexandre Sablayrolles, and Herv{\'e} J{\'e}gou.
\newblock Training data-efficient image transformers \& distillation through attention.
\newblock In \emph{International Conference on Machine Learning}, pages 10347--10357, 2021.

\bibitem[Turner(2015)]{Turner2015}
Jefferson~T. Turner.
\newblock Zooplankton fecal pellets, marine snow, phytodetritus and the ocean’s biological pump.
\newblock \emph{Progress in Oceanography}, 130:\penalty0 205--248, 2015.

\bibitem[van~den Oord et~al.(2015)van~den Oord, Korshunova, Burms, Degrave, Pigou, Buteneers, and Dieleman]{deepsea2015}
Aäron van~den Oord, Ira Korshunova, Jeroen Burms, Jonas Degrave, Lionel Pigou, Pieter Buteneers, and Sander Dieleman.
\newblock Classifying plankton with deep neural networks.
\newblock https://sander.ai/2015/03/17/plankton.html, 2015.
\newblock [Online; accessed February, 24, 2025].

\bibitem[Wightman(2019)]{rw2019timm}
Ross Wightman.
\newblock Pytorch image models.
\newblock \url{https://github.com/rwightman/pytorch-image-models}, 2019.

\bibitem[Xie et~al.(2022)Xie, Zhang, Cao, Lin, Bao, Yao, Dai, and Hu]{xie2022simmim}
Zhenda Xie, Zheng Zhang, Yue Cao, Yutong Lin, Jianmin Bao, Zhuliang Yao, Qi Dai, and Han Hu.
\newblock {SimMIM}: A simple framework for masked image modeling.
\newblock In \emph{CVPR}, pages 9653--9663, 2022.

\bibitem[Zbontar et~al.(2021)Zbontar, Jing, Misra, LeCun, and Deny]{zbontar2021barlow}
Jure Zbontar, Li Jing, Ishan Misra, Yann LeCun, and St{\'e}phane Deny.
\newblock Barlow twins: Self-supervised learning via redundancy reduction.
\newblock In \emph{International Conference on Machine Learning}, pages 12310--12320, 2021.

\end{thebibliography}
